\documentclass{article}

\usepackage[english]{babel}
\usepackage{kotex}
\usepackage[unicode, colorlinks, urlcolor=cyan]{hyperref}
\usepackage{txfonts}
\usepackage{CJKutf8}

\usepackage[letterpaper,top=2cm,bottom=2cm,left=3cm,right=3cm,marginparwidth=1.75cm]{geometry}

\usepackage{graphicx}
\usepackage{pgfplots}
\usepackage{multirow}
\usepackage{multicol}
\usepackage{caption}
\usepackage{subcaption}
\usepackage{subfiles}
\usepackage{xcolor, colortbl}
\usepackage[toc,page]{appendix}

\usepackage[T1]{fontenc}
\usepackage{CJKutf8}

\makeatletter

\newcommand{\printfnsymbol}[1]{%
  \textsuperscript{\@fnsymbol{#1}}%
}
\newcommand{\jcomma}{\hspace{-.5em}$_{_{、}}$\hspace{.8em}}
\newcommand{\jperiod}{\hspace{-.5em}$_{_{。}}$\hspace{.5em}}

\title{JaQuAD: Japanese Question Answering Dataset \\ for Machine Reading Comprehension}
\author{
    ByungHoon So\thanks{Equal contribution}\qquad
    Kyuhong Byun\printfnsymbol{1}\qquad
    Kyungwon Kang\qquad
    Seongjin Cho\\
    \texttt{\{byunghoon, khbyun, kangnak, sjcho\}@skelterlabs.com}
}
\date{}

\pgfplotsset{compat=1.17}
\begin{document}

\maketitle

\begin{abstract}
Question Answering (QA) is a task in which a machine understands a given document and a question to find an answer.
Despite impressive progress in the NLP area, QA is still a challenging problem, especially for non-English languages due to the lack of annotated datasets.
In this paper, we present the \textbf{Ja}panese \textbf{Qu}estion \textbf{A}nswering \textbf{D}ataset, JaQuAD, which is annotated by humans.
JaQuAD consists of 39,696 extractive question-answer pairs on Japanese Wikipedia articles.
We finetuned a baseline model which achieves 78.92\% for F1 score and 63.38\% for EM on test set.
The dataset and our experiments are available at \texttt{https://github.com/SkelterLabsInc/JaQuAD}.
\end{abstract}

\section{Introduction}
    \label{chap:Introduction}

Question Answering (QA), a.k.a. Reading Comprehension (RC), is a natural language processing task in which a machine understands a given document and a question to find an answer.
This task has become popular with the emergence of a large-scale and high-quality QA dataset named SQuAD~\cite{SQuAD1, SQuAD2}, leading to the release of other datasets such as Natural Questions~\cite{NaturalQuestions}, CoQA~\cite{CoQA}, and HotpotQA~\cite{HotpotQA}.
These datasets contributed impressive progress for English question answering models over the past few years.

Naturally, a variety of studies have emerged for non-English QA.
Some researchers made substantial efforts to construct non-English QA datasets.
Similar to SQuAD, large-scale and high-quality datasets have been proposed, such as KorQuAD in Korean~\cite{KorQuAD1, KorQuAD2}, FQuAD in French~\cite{FQuAD}, and GermanQuAD in German~\cite{GermanQuAD}.
As a more general solution for non-English QA, other studies proposed multilingual models~\cite{Lewis19, XLM-R} or techniques transferring a monolingual model to the target language~\cite{Artetxe20}.
Although these approaches solved QA without training data of the target language, the reported performance failed to achieve comparable results compared to the performance training with the target language data~\cite{KorQuAD2, FQuAD, GermanQuAD}.

To fill the gap for the Japanese language, we propose a Japanese Question Answering Dataset (JaQuAD) to address the need for a large-scale and high-quality Japanese QA dataset.
JaQuAD contains 39,696 question-answer pairs annotated by humans and 12,348 contexts which span over one or more paragraphs from 901 Japanese Wikipedia articles.
More specifically, the training, development, and test sets of JaQuAD contain 31,748, 3,939, and 4,009 question-answer pairs, respectively.

To evaluate the JaQuAD, we finetuned BERT-Japanese~\cite{BertJapanese}, a transformer-based pretrained language model, on JaQuAD.
This baseline achieves 78.92\% for F1 score and 63.38\% for EM on test set.
It suggests there is plenty of room for improvement in modeling and learning on the JaQuAD dataset.

\section{Related work}

\subsection{Reading comprehension in English}
The Reading Comprehension~\cite{MCTest, SQuAD1} attempts to solve the Question Answering problem by finding the span in one or several paragraphs which answer a given question.
In recent years, English Question Answering models made impressive progress.
This progress was affected by the release of large and realistic English QA datasets such as SQuAD~1.1~\cite{SQuAD1}, SQuAD~2.0~\cite{SQuAD2}, MS Marco~\cite{MSMarco}, Natural Questions~\cite{NaturalQuestions}, QuAC~\cite{QuAC}, CoQA~\cite{CoQA}, and HotpotQA~\cite{HotpotQA}.
Among them, SQuAD~1.1 has been one of the most famous reference datasets which consists of Wikipedia documents and question-answer pairs generated by crowd workers.
Based on SQuAD~1.1, each of the datasets introduced its subtleties.
SQuAD~2.0 introduced unanswerable adversarial questions.
MS Marco and Natural Questions contain much larger questions by using Bing and Google search logs, respectively, rather than generated by crowd workers.
CoQA and QuAC were built for conversational QA (multi-turn QA) and free-form answers.
HotpotQA introduced multi-hop questions where the answer must be found among multiple documents.

\subsection{Reading comprehension in other languages}
There were a few SQuAD format datasets released in non-English languages.
Some examples are KorQuAD 1.0~\cite{KorQuAD1}, KorQuAD 2.0~\cite{KorQuAD2}, KLUE-MRC~\cite{KLUE-MRC}, FQuAD 1.1~\cite{FQuAD}, GermanQuAD~\cite{GermanQuAD}, and SberQuAD~\cite{SberQuAD}.
KorQuAD~1.0 is a Korean QA dataset that contains over 70k samples.
KorQuAD~2.0 is another Korean QA dataset that contains over 100k samples whose contexts are HTML contents from Korean Wikipedia, not plain text contents.
KLUE-MRC is a Korean QA dataset that contains over 29k samples which includes unanswerable questions and plausible fake answers.
FQuAD~1.1 is a French QA dataset that contains over 60k samples.
GermanQuAD is a German QA dataset that contains over 13k samples.
SberQuAD is a Russian QA dataset that contains over 50k samples.
Table~\ref{tab:Benchmarks} lists some of the available datasets along with the number of question-answer pairs they contain~\cite{NLPProgress}.
For comparison, Table~\ref{tab:Benchmarks} also includes some English QA datasets and JaQuAD.

\begin{table}[t]
\centering
\begin{tabular}{lcl}
    Dataset & Language  & Size    \\ \hline
    SQuAD 1.1 & English & 100k+ \\
    SQuAD 2.0 & English & 150k \\
    MS Marco & English & 100k \\
    CoQA & English & 127k+ \\
    HotpotQA & English & 113k+ \\ \hline
    KorQuAD 1.0 & Korean & 70k+ \\
    KorQuAD 2.0 & Korean & 100k+ \\
    KLUE-MRC & Korean & 29k+ \\ \hline
    FQuAD 1.1 & French & 62k+ \\
    GermanQuAD & German & 13k+ \\
    SberQuAD & Russian & 50k+ \\ \hline
    JaQuAD & Japanese & 39k+ \\ \hline
\end{tabular}
\caption {Samples of existing QA datasets}
\label{tab:Benchmarks}
\end{table}

In the case of Japanese, a few Japanese QA datasets were released such as a driving-domain RC-QA dataset~\cite{JaDrivingQAData} and an answerability annotated RC dataset~\cite{JaEqidenQAData}.
The driving-domain RC-QA dataset contains over 20k samples.
However, the documents of the dataset come from driving-blogs which limits its domain and patterns.
The answerability annotated RC dataset contains 12k question-answer pairs over 56k contexts.
The question-answer pairs come from buzzer quizzes of the abc/EQIDEN competition and the contexts are paragraphs automatically collected from Wikipedia articles.
Later, the answerability score of a context is annotated by crowdsourcing.
Thus, this dataset has a shortcoming that the majority of contexts and corresponding questions are contrived, and that they could be biased by the original competition in some ways.

\section{Dataset Collection}
Referencing the data collection of SQuAD 1.1, we collect our dataset in three stages: (1) collecting contexts, (2) generating question-answer pairs on those contexts, and (3) validating collected questions and answers.
For stages (2) and (3), human annotators were selected through a qualification test.
We asked annotators to generate question-answer pairs from a Wikipedia document and news articles as the qualification test.
Then, only the annotators who generated fluent questions and produced answers in a consistent format participated in dataset collection.

\subsection{Contexts collection}
We chose Japanese Wikipedia pages as contexts to cover a wide range of domains.
We collected 799 articles from the Japanese Wikipedia pages referencing quality articles~\cite{WikipediaJaGoodArticles,  WikipediaJaFeaturedArticles}.
Also, we collected 102 articles in other categories to broaden the domains of our dataset.
901 articles are randomly split to training, development, and test sets which are made up of 691, 101, and 109 articles, respectively.

From each article, we extracted a context consisting of an individual paragraph or consecutive paragraphs without an image, a figure, and a table.
As a result, training, development, and test sets are made up of 9,713, 1,431, and 1,479 contexts, respectively.

\subsection{Question and answer pairs generation}
Human annotators generated multiple question-answer pairs while reading a context.
A question had to be answerable using the content of the corresponding context.
And an answer must be a span in the context.
For the consistency of the answers, all annotators generated answers according to the criteria provided.
For example, the answer should be the minimum span corresponding to the question and the basic unit should be included when asked about the number.
The details are in Appendix A.
Also, we asked annotators to tag the answer types and question types used in KorQuAD 1.0 and gave feedback on balancing the question and answer type to control the distribution of each type.
As a result, 39,696 question-answer pairs are generated.
Training, development, and test sets have 31,748, 3,939, and 4,009 question-answer pairs, respectively.

\subsection{Quality management}
For credible evaluation, collected data went through a cross-validation process.
During this process, annotators validated not only questions and answers, but answer types and question types.
While validating the answer and question types, validators double-checked the logical process of inferring answers.
Thus, they naturally validate the answerability of the question in-depth.
Ambiguous question-answer pairs were corrected through discussion in the annotator group.
During validation, we fixed either question or answer in 4693 question-answer pairs and the question type or answer type of 1807 question-answer pairs.

\section{Dataset Analysis}
    \label{chap:DatasetAnalysis}
This chapter compares JaQuAD with two datasets in the same format, SQuAD~1.1 and KorQuAD 1.0.
Referencing the analyses on SQuAD~1.1~\cite{Liu18} and KorQuAD 1.0~\cite{KorQuAD1}, we analyzed questions and answers of JaQuAD.
First, we categorized answers by predefined types and measured the distribution of each answer type.
Second, we categorized questions according to the required reasoning ability and measured the distribution of each question type.
Finally, we measured the distribution of the lengths of contexts and answers.

\subsection{Answer type analysis}
    \label{chap:AnswerTypeAnalysis}
\begin{table}[t]
\centering
\begin{tabular}{l|rrrr}
    Answer Types & \multirow{2}{*}{SQuAD~1.1} & SQuAD~1.1 & \multirow{2}{*}{KorQuAD~1.0} & \multirow{2}{*}{JaQuAD} \\
    (Question interrogatives) & & except Others type  & &  \\ \hline
    Object (What/Which) & 49.4\% & 60.3\% & 55.4\% & 49.4\% \\
    Person (Who) & 10.0\% & 12.2\% & 23.2\% & 15.5\% \\
    Date/Time (When) & 6.6\% & 8.1\% & 8.9\% & 19.5\% \\
    Location (Where) & 4.1\% & 5.0\% & 7.5\% & 14.1\% \\
    Manner (How) & 10.3\% & 12.6\% & 4.3\% & 0.5\% \\
    Cause (Why) & 1.5\% & 1.8\% & 0.7\% & 1.0\% \\
    Others & 18.1\% & - & - & 0.0\% \\ \hline
\end{tabular}
\caption {Distribution of answer types. For JaQuAD, the distribution was extracted from the entire dataset. For KorQuAD~1.0 and SQuAD~1.1, the distributions are extracted from the random 280 and 192 samples from the development set, respectively. ``SQuAD~1.1 except Others type'' represents the proportions of the remaining types calculated by taking the remaining 81.9\% as the total, excluding the Others type.}
\label{tab:AnswerTypesRatio}
\end{table}

\begin{CJK}{UTF8}{min}
\begin{table}[!ht]
\centering
\begin{tabular}{l}
    \hline
    \cellcolor{gray!15} Syntactic variation \\ \hline
    Question: 心臓疾患の原因とされているのは何ですか? (What is the cause of heart disease?) \\ \hline
    血中の\underline{酸化型LDL}コレステロールは心臓疾患の原因になると考えられ\jcomma\ldots \\
    (\underline{Oxidized LDL} cholesterol in the blood is thought to cause heart disease, \ldots) \\ \hline
    
    \cellcolor{gray!15} Lexical variation (synonymy) \\ \hline
    Question: アンネの日記に書かれている期間はどれくらい? (How long has Anne's Diary been written?) \\\hline
    ここでの生活は\underline{2年間}に及び\jcommaその間\jcommaアンネは隠れ家でのことを日記に書き続けた\jperiod \\
    (Life here lasted \underline{for two years}, in the meantime, Anne kept writing about her hideout in her diary.) \\ \hline
    \cellcolor{gray!15} Lexical variation (world knowledge) \\ \hline
    Question: デング熱の媒介者は何ですか\jperiod(What are the mediators of dengue fever?)  \\\hline
    デング熱は\underline{蚊}の吸血活動を通じて\jcommaウイルスが人から人へ移り\jcomma高熱に達することで知られる一過\\性の熱性疾患であり\jcomma\ldots \\
    (Dengue fever is a transient febrile disease known to transfer the virus from person to person and reach high\\ fever through the bloodsucking activity of \underline{mosquitoes}, \ldots) \\ \hline
    
    \cellcolor{gray!15} Multiple sentence reasoning \\ \hline
    Question: 直接押出と間接押出では\jcomma一般的にどちらの技法の方がより大きな力を必要としますか? \\
    (Among direct extrusion or indirect extrusion, which technique generally requires more force?) \\\hline
    \underline{直接押出}または前方押出は\jcomma最も一般的な押出しプロセスである\jperiod \ldots 全行程で周囲の壁との間に摩\\擦を生じるため\jcomma一般に間接押出よりも大きな力を必要とする\jperiod \ldots \\
    (\underline{Direct extrusion} or forward extrusion is the most common extrusion process. \ldots it generally requires more\\ force than indirect extrusion because it creates friction with the surrounding walls during the entire process\ldots) \\ \hline
    
    \cellcolor{gray!15} Logical reasoning \\ \hline
    Question: 第1国会の選挙でどんな１票の影響力が最も小さかったのは\jcommaどんな人々だったの?\\(What kind of people had the least influence on one vote in the first parliamentary elections?) \\\hline
    \ldots 土地所有者の1票が都市民2票·民15票·\underline{労働者}45票に相当するという極めて不平等な選挙制度で\\あった\jperiod\ldots\\
    (\ldots It was an extremely unequal election system in which 1 vote for landowners was equivalent to 2 votes for\\ citizens, 15 votes for farmers, and 45 votes for \underline{workers}. \ldots) \\ \hline
\end{tabular}
\caption {An example for each question type. The answers are underlined.}
\label{tab:QuestionTypesExample}
\end{table}
\end{CJK}

The first analysis aims to understand the answer types of the dataset.
For the answer type analysis, annotators manually labeled the answer types during data collection and validation.
We used six answer types used in the analysis of KorQuAD~1.0: Object, Person, Date/Time, Location, Manner, and Cause.
Further, we compared JaQuAD with SQuAD~1.1 by matching answer types to corresponding question types of SQuAD~1.1 which are categorized based on interrogative~\cite{Liu18}.
Among them, the “Others” question type is removed, and “What” and “Which” question types are merged into the “Object” answer type because they are hard to distinguish clearly by answers.

Table~\ref{tab:AnswerTypesRatio} shows the distribution of answer types on SQuAD 1.1, KorQuAD 1.0, and JaQuAD.
In the Answer Types column, corresponding question types used in SQuAD~1.1 are described in parentheses.
Because KorQuAD 1.0 and JaQuAD do not have Others type, we added the column “SQuAD~1.1 except Others type” for direct comparison.
This column represents the proportions of the remaining types calculated by taking the remaining 81.9\% as the total, excluding the Others type.
In SQuAD~1.1 and KorQuAD~1.0, Object type occupies more than half of the dataset.
Similarly, Object type occupies the largest proportion in JaQuAD, but it decreases to 49.4\%.
Compared to the other datasets, the proportion of Date/Time and Location types significantly increased to 19.7\% and 14.0\%, respectively.
The proportion of Person type is 15.4\%, similar to that of the SQuAD~1.1.
As a result, the proportions of the Person, Date/Time, and Location types are similar.
In contrast, the proportion of Manner and Cause types significantly decreases to 0.5\% and 1.0\%, respectively.

\subsection{Question type analysis}
    \label{chap:QuestionTypeAnalysis}

The second analysis aims to understand the question types of the dataset.
Referencing KorQuAD 1.0, we categorized questions into five question types according to the reasoning ability required to answer the question.
Table~\ref{tab:QuestionTypesExample} shows an example for each question type.
\textit{Syntactic variation} implies that a question is made by changing the word order or reorganizing the sentence.
\textit{Lexical variation (synonymy)} implies that the keywords in a question are transformed to a synonym compared to the context.
\textit{Lexical variation (world knowledge)} implies that world knowledge is required to match keywords in a question and the corresponding words in the context.
\textit{Multiple sentence reasoning} implies that reasoning over multiple sentences is required to answer the question.
\textit{Logical reasoning} implies that the question requires multi-hop reasoning to find the answer among the multiple options in the context: the answer could be found by matching the condition of the question or actively using the information in parentheses.
This type of question often requires comparing features of listed items or using implicit information in the context.

\begin{table}[t]
\centering
\begin{tabular}{l|rrr}
    Types & SQuAD~1.1 & KorQuAD~1.0 & JaQuAD \\ \hline
    Syntactic variation & 64.1\% & 56.4\% & 32.0\% \\
    Lexical variation (synonymy) & 33.3\% & 13.6\% & 16.7\% \\
    Lexical variation (world knowledge) & 9.1\% & 3.9\% & 21.2\% \\
    Multiple sentence reasoning & 13.6\% & 19.6\% & 20.2\% \\
    Logical reasoning & - & 3.6\% & 10.0\% \\
    Ambiguous & 6.1\% & 2.9\% & - \\ \hline
\end{tabular}
\caption {Distribution of question types. For JaQuAD, the distribution was extracted from the entire dataset. For KorQuAD~1.0 and SQuAD~1.1, we use the values in their paper which are the results of the random 280 and 192 samples from the development set, respectively.}
\label{tab:QuestionTypesRatio}
\end{table}

Table~\ref{tab:QuestionTypesRatio} shows the proportion of question types of SQuAD~1.1, KorQuAD~1.0, and JaQuAD.
The Ambiguous type includes the cases that authors of the paper disagree with annotators’ answers, a question does not have a unique answer, and external knowledge out of the corresponding context is needed to answer a question.

The proportion of Syntactic variation type is almost half compared to SQuAD~1.1 and KorQuAD~1.0.
In contrast, the proportions of Lexical variation (world knowledge) and Logical reasoning types are more than double.
The proportion of Lexical variation (synonymy) and Multiple sentence reasoning types remains similar to KorQuAD~1.0.
These represent JaQuAD requires a higher level of reasoning ability than SQuAD~1.1 and KorQuAD~1.0.

\subsection{Context and answer length}
    \label{chap:ContextAndAnswerLength}

\begin{figure}[t]
\centering
\begin{subfigure}[]{.45\textwidth}
    \centering
    \scalebox{1.}{
    \begin{tikzpicture}
    \begin{axis}[
         ylabel near ticks,
        ymin=0, ymax=13000,
        enlargelimits=false,
        xlabel=\# tokens,
        scaled ticks={base 10:0},
        ]
    \addplot
    	[const plot,fill=blue!25,draw=black!70]
    coordinates
    {
(0, 2850)
(100, 11365)
(200, 12089)
(300, 7527)
(400, 3719)
(500, 1123)
(600, 481)
(700, 266)
(800, 262)
(900, 5)
(1000, 6)
(1100, 3)} 
    	\closedcycle;
    \end{axis}
    \end{tikzpicture}
    }
    \caption{Distribution of context lengths}
    \label{fig:ContextLengths}
\end{subfigure}
\hfill
\begin{subfigure}[]{.45\textwidth}
    \centering
    \scalebox{1.}{
    \begin{tikzpicture}
    \begin{axis}[
         ylabel near ticks,
        ymin=0, ymax=21000,
        enlargelimits=false,
        xlabel=\# tokens,
        scaled ticks={base 10:0},
        ]
    \addplot
    	[const plot,fill=blue!25,draw=black!70]
    coordinates
    {
(0, 1800)
(10, 20125)
(20, 13063)
(30, 3592)
(40, 816)
(50, 196)
(60, 71)
(70, 18)
(80, 6)
(90, 2)
(100, 2)
(110, 4)
(120, 1)} 
    	\closedcycle;
    \end{axis}
    \end{tikzpicture}
    }
    \caption{Distribution of question lengths}
    \label{fig:QuestionLengths}
\end{subfigure}
\hfill
\begin{subfigure}[]{.45\textwidth}
    \centering
    \scalebox{1.}{
    \begin{tikzpicture}
    \begin{axis}[
         ylabel near ticks,
        ymin=0, ymax=20000,
        enlargelimits=false,
        xlabel=\# tokens,
        scaled ticks={base 10:0},
        ]
    \addplot
    	[const plot,fill=blue!25,draw=black!70]
    coordinates
    {
(0, 17570)
(2, 13916)
(4, 5451)
(6, 1746)
(8, 591)
(10, 188)
(12, 90)
(14, 56)
(16, 38)
(18, 14)
(20, 8)
(22, 6)
(24, 5)
(28, 4)} 
    	\closedcycle;
    \end{axis}
    \end{tikzpicture}
    }
    \caption{Distribution of answer lengths}
    \label{fig:AnswerLengths}
\end{subfigure}
\hfill
\begin{subfigure}[]{0.45\textwidth}\end{subfigure}
\caption{Distribution of context, question, and answer lengths.}
\label{fig:ContextAnswerLengths}
\end{figure}
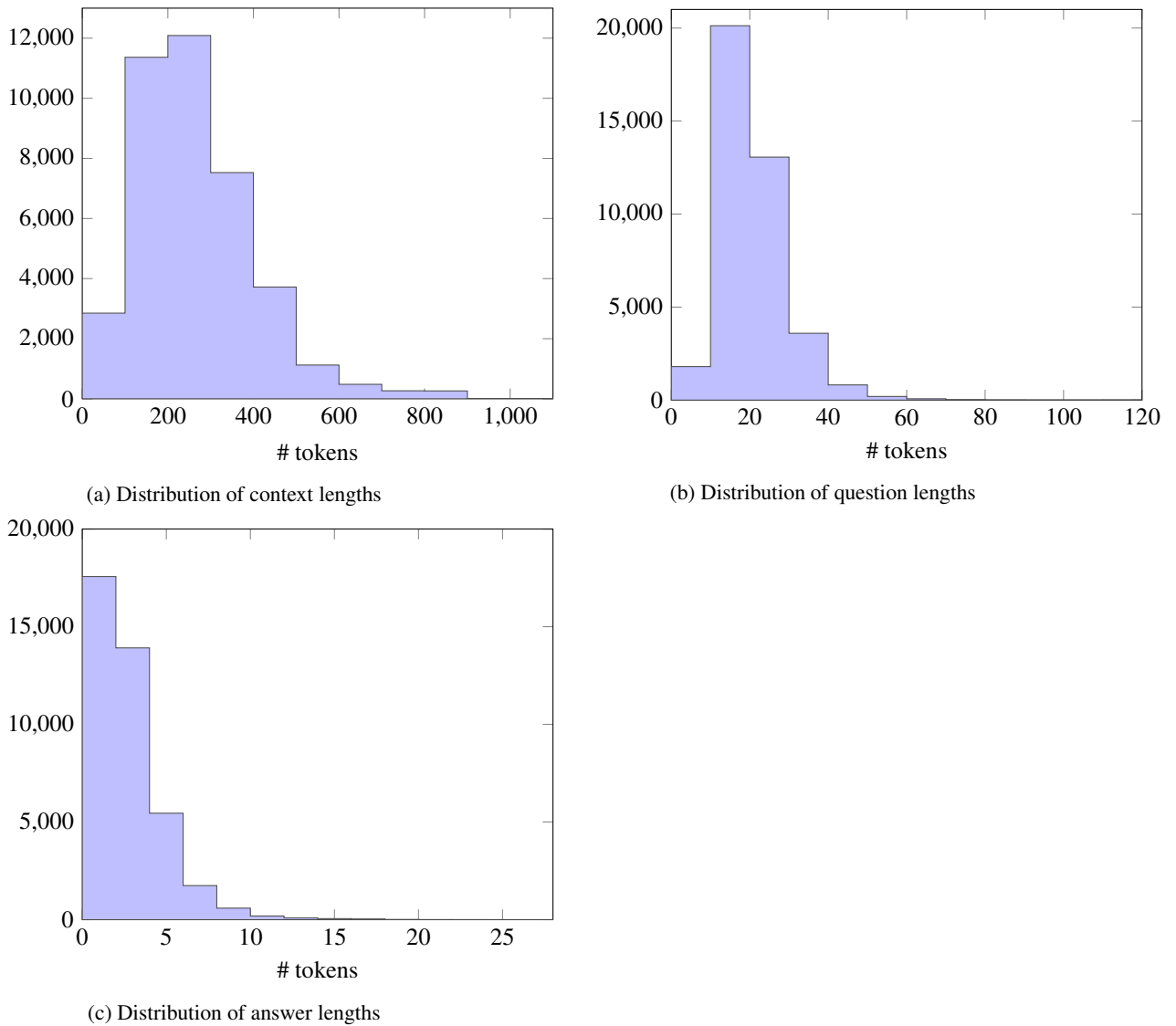

The third analysis aims to understand the distribution of context and answer lengths.
The length represents the number of tokens.
We use the tokenizer of BERT-Japanese in the transformers library~\cite{HuggingFace}.
The tokenizer first segments a text into morphemes using MeCab tokenzier~\cite{MeCab} and the Unidic 2.1.2 dictionary~\cite{Unidic}.
Then, the tokenizer generates subword tokens from each morpheme.
The vocabulary size of BERT-Japanese is 32,000.

Figure~\ref{fig:ContextAnswerLengths} shows the distribution of context lengths, question lengths, and answer lengths.
In Figure~\ref{fig:ContextLengths}, the context length ranges from 22 to 1,130, and the average is 266.
5.0\% of contexts are longer than 500 tokens.
As the max sequence length of pre-trained models is 512 in general, truncation of these contexts could affect the performance of a model.
In Figure~\ref{fig:QuestionLengths}, the question length ranges from 5 to 126, and the average is 21.1.
As all of the questions are shorter than 128 tokens, we did not truncate a question during training.
In Figure~\ref{fig:AnswerLengths}, the answer length ranges from 1 to 70, and the average is 3.3.
Most of the answers are short, and only 4.1\% are longer than 8 tokens.
Therefore, JaQuAD has lower coverage for long answers compared to other datasets.

\section{Dataset Evaluation}
We use two evaluation metrics, Exact Match (EM) and F1 score, which are common metrics for evaluating performances of QA models.

\noindent\textbf{Exact Match (EM).} This metric measures the percentage of predictions that match exactly to the ground truth answer.

\noindent\textbf{F1 score.} This metric measures the overlap between the prediction and the ground truth answer.
We treat the prediction and the ground truth as bags of characters, compute F1 score, and average over all of the questions.
Note that although SQuAD used word-level F1 score, we computed character-level F1 score.

\begin{CJK}{UTF8}{min}
\begin{table}[t]
    \centering
    \begin{tabular}{|l|}
        \hline
        Context: ヒトラーは\jcomma周辺各国のドイツ系住民処遇問題に対し民族自決主義を主張し\jcommaドイツ人居\\住地域のドイツへの併合を要求した\jperiod\ldots \\
        (Hitler insisted on self-determination for the treatment problem of German residents in neighboring countries\\ and demanded the merger of German settlements with Germany. \ldots)\\ \hline
        Question: ヒトラーは\jcomma周辺各国のどのような問題に対し民族自決主義を主張し\jcommaドイツ人居住地域\\のドイツへの併合を要求したか？\\(For what problems did Hitler insist on self-determination and demand the annexation of German settlements\\ into Germany?)\\\hline
        Ground Truth: ドイツ系住\textcolor{blue}{民}\textcolor{red}{処遇問題} \\ 
        English translation: \textcolor{red}{treatment problem} of German residents \\ \hline
        Prediction: \textcolor{red}{処遇問題}に対し\textcolor{blue}{民}族自決主義 \\ 
        English translation: self-determination for \textcolor{red}{treatment problem} \\ \hline
        Character-level F1 in Japanese: 43.5\% \\
        Word-level F1 in English: 44.4\% \\ \hline
    \end{tabular}
    \caption{An example of calculating F1 scores in English and Japanese.}
    \label{tab:ComputingF1}
\end{table}

Computing F1 in words is not trivial in Japanese because Japanese sentences do not have spaces.
We chose a character-level F1 score as an evaluation metric by referring to the use of character-based evaluation metrics in Korean QA datasets~\cite{KorQuAD1, KorQuAD2, KLUE-MRC}.
Because Japanese uses thousands of kanji (Chinese characters) and each kanji has a meaning, the probability of two phrases coincidentally overlapping by character is low when the two phrases have different meanings.
Table~\ref{tab:ComputingF1} shows an example of calculating the character-level F1 score in Japanese and the word-level F1 scores in English.
When we translate the ground truth and prediction in English, ``treatment problem of German residents'' (イツ系住民処遇問題) and ``self-determination for treatment problem'' (処遇問題に対し民族自決主義) overlap two words in English (`treatment' and `problem').
No words overlap coincidentally.
The comparison in Japanese is similar; ドイツ系住民処遇問題 (treatment problem of German residents) and 処遇問題に対し民族自決主義 (self-determination for treatment problem) overlap five characters (処, 遇, 問, 題, and 民).
Only one Japanese character (民), which means people or nation, overlap coincidentally.

Although the evaluation process in SQuAD ignores punctuation and articles (a, an, the), we did not need them because Japanese has no articles, and there is no punctuation in the ground truth answers of JaQuAD.
Thus, we could remain consistent with the former approach without any additional filtering.
\end{CJK}

\section{Experiments}

\subsection{Experimental setup}
The baseline model is a Japanese pre-trained language model, BERT-Japanese~\cite{BertJapanese}, published in HugginFace’s transformers library~\cite{HuggingFace}.
We trained the baseline model on JaQuAD for four epochs with a learning rate of $2·10^{-5}$ using AdamW with default settings.
The learning rate is linearly increased for the first 10\% of steps and linearly decreased to zero afterward.
The batch size is set to 32 and a maximum sequence length of 384 tokens.
We did not truncate a question, because the longest question has 126 tokens.
However, contexts could be truncated to meet the maximum sequence length.
All the experiments were carried out with the HuggingFace transformers library~\cite{HuggingFace} and trained with cloud TPUs provided by TPU Research Cloud program~\footnote{https://sites.research.google/trc/}.

\subsection{Model performance}
Table~\ref{tab:BaselinePerformance} shows the performance of the baseline model on the development and the test sets of JaQuAD.
The baseline achieves 78.92\% for F1 score and 63.38\% for EM on test set.
Further, we analyzed the performance of the baseline by the question types, answer types, and answer lengths described in Chapter~\ref{chap:DatasetAnalysis}.
\begin{table}[t]
\centering
\begin{tabular}{llrrrr}
    \multirow{2}{*}{Model} & \multirow{2}{*}{Model size} & \multicolumn{2}{c}{Development} & \multicolumn{2}{c}{Test} \\    % BERT-Base
    % \multirow{2}{*}{Model}& \multicolumn{2}{c}{Development} & \multicolumn{2}{c}{Test} \\   
    & & F1 & EM & F1 & EM \\ \hline
    BERT-Japanese & BERT$_{BASE}$ & 77.35 & 61.01 & 78.92 & 63.38 \\ \hline
\end{tabular}
\caption {Performance of the baseline on JaQuAD}
\label{tab:BaselinePerformance}
\end{table}

\subsubsection{Performance by answer types}
In order to understand the effect of answer types on model performance, we analyzed the model performance according to the answer type.
We have explored the answer types and the distribution of them in Chapter~\ref{chap:AnswerTypeAnalysis}.
Figure~\ref{fig:PerformanceAnswerType} shows the F1 and EM scores of the baseline for each answer type.
The model shows much better performance on Date/Time than the average, followed by Person, Object, and Location types.
The model seems to perform better on types with few plausible candidates, such as Date/Time and Person types.
Note that the proportions of Manner and Cause types are up to 1\%.
These types seem to suffer from the lack of training data.

\subsubsection{Performance by question types}
\begin{figure}[t]
\centering
\begin{subfigure}[]{.45\textwidth}
    \scalebox{.85}{
    \begin{tikzpicture}
    \begin{axis}[
        ybar, ylabel near ticks,
        enlargelimits=0.15,
        legend style={at={(0.5,-0.25)},
          anchor=north,legend columns=-1},
        ylabel={Performance(\%)},
        symbolic x coords={
        Object, Person, Date/Time, Location, 
        Manner, Cause,
        Avg.},
        xtick=data,
        nodes near coords,
        nodes near coords align={vertical},
        x tick label style={rotate=45,anchor=east},
        bar width=7pt,
        ]
    \addplot coordinates {
    (Object,74.99) (Person,78.84) (Date/Time,87.91) (Location,72.51) 
    (Manner,64.75) (Cause,52.10) 
    (Avg.,77.35)};
    \addplot coordinates {
    (Object,58.99) (Person,63.14) (Date/Time,69.77) (Location,56.73) 
    (Manner,41.18) (Cause,34.04) 
    (Avg.,61.01)};
    \legend{F1 score, EM}
    \end{axis}
    \end{tikzpicture}
    }
    \caption{Performance for each answer type.}
    \label{fig:PerformanceAnswerType}
\end{subfigure}
\hfill
\begin{subfigure}[]{.45\textwidth}
    \scalebox{.85}{
    \begin{tikzpicture}
    \begin{axis}[
        ybar, ylabel near ticks,
        enlargelimits=0.15,
        legend style={at={(0.5,-0.25)},
          anchor=north,legend columns=-1},
        ylabel={Performance(\%)},
        symbolic x coords={
        % Syntactic variation, 
        % Lexical variation (synonymy), 
        % Lexical variation (word knowledge), 
        % Multiple sentence reasoning, 
        % Logical reasoning, 
        Syn., 
        Lex. (syn), 
        Lex. (world), 
        Mul. sent., 
        Logical, 
        Avg.},
        xtick=data,
        nodes near coords,
        nodes near coords align={vertical},
        x tick label style={rotate=45,anchor=east},
        bar width=7pt,
        ]
    \addplot coordinates {
    % (Syntactic variation, 70)
    % (Lexical variation (synonymy), 70)
    % (Lexical variation (word knowledge), 70)
    % (Multiple sentence reasoning, 70)
    % (Logical reasoning, 70)
    (Syn., 87.06)
    (Lex. (syn), 83.58)
    (Lex. (world), 75.37)
    (Mul. sent., 76.43)
    (Logical, 53.71)
    (Avg., 77.35)};
    \addplot coordinates {
    % (Syntactic variation, 70)
    % (Lexical variation (synonymy), 70)
    % (Lexical variation (word knowledge), 70)
    % (Multiple sentence reasoning, 70)
    % (Logical reasoning, 70)
    (Syn., 73.64)
    (Lex. (syn), 67.86)
    (Lex. (world), 57.74)
    (Mul. sent., 60.07)
    (Logical, 33.26)
    (Avg., 61.01)};
    \legend{F1 score, EM}
    \end{axis}
    \end{tikzpicture}
    }
    \caption{Performance for each question type.}
    \label{fig:PerformanceQuestionType}
\end{subfigure}
\caption{Performance for answer type and question type.}
\end{figure}
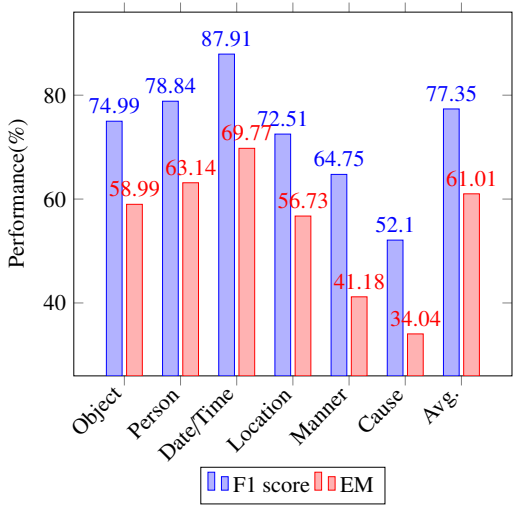
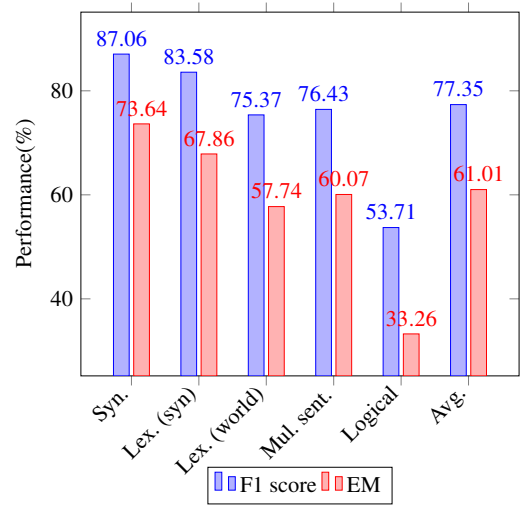

In order to understand the effect of required reasoning ability on model performance, we analyzed the model performance according to the question type.
We have explored the question types and the distribution of them in Chapter~\ref{chap:QuestionTypeAnalysis}.
Figure~\ref{fig:PerformanceQuestionType} shows the F1 and EM scores of the baseline for each question type.
Syn., Lex. (syn), Lex. (world), Mul. sent., and Logical represent Syntactic variation, Lexical variation (synonymy), Lexical variation (world knowledge), Multiple sentence reasoning, and Logical reasoning types, respectively.
The model performed best on Syntactic variation type and performed the second-best on Lexical variation (synonymy) type.
The performances on Lexical variation (world knowledge) and Multiple sentence reasoning are similar to the average.
As expected, the performance on Logical reasoning type is the lowest of all types, which is more than 20\%p lower than the average.

\subsubsection{Performance by answer lengths}

\begin{figure}[h!]
\centering
\begin{tikzpicture}
\begin{axis}[
    ybar, ylabel near ticks,
    enlargelimits=0.15,
    legend style={at={(0.5,-0.25)},
      anchor=north,legend columns=-1},
    ylabel={Performance(\%)},
    symbolic x coords={1-2, 3-4, 5-6, 7-8, 9+, Avg.
    },
    xtick=data,
    nodes near coords,
    nodes near coords align={vertical},
    bar width=7pt,
    ]
\addplot coordinates {
(1-2, 73.36)
(3-4, 81.33)
(5-6, 78.29)
(7-8, 81.23)
(9+, 70.12)
(Avg., 77.35)};
\addplot coordinates {
(1-2, 59.52)
(3-4, 65.04)
(5-6, 57.74)
(7-8, 56.55)
(9+, 46.94)
(Avg., 61.01)};
\legend{F1 score, EM}
\end{axis}
\end{tikzpicture}\caption{Performance by answer length.}
\label{fig:PerformanceAnswerLength}
\end{figure}
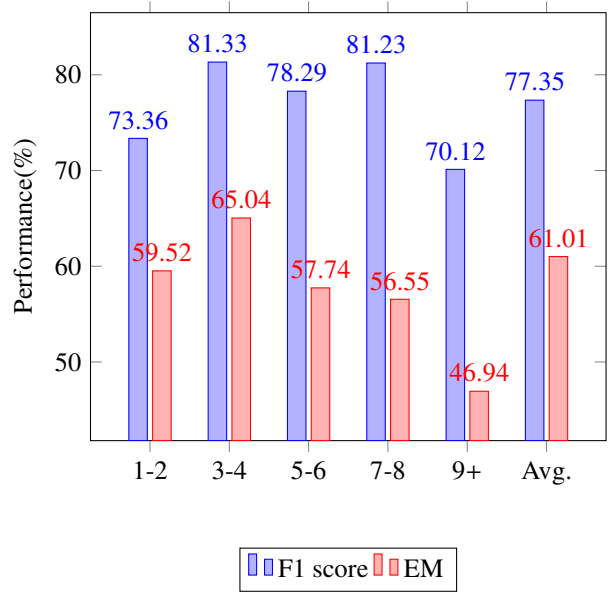

We analyzed the model performance according to the answer lengths.
Figure~\ref{fig:PerformanceAnswerLength} shows the F1 and EM scores of the baseline according to the answer lengths.
The model performed best on the answers with length 3-4 tokens, but the f1 score are similar when the answer lengths are 3-8 tokens.
The model shows the lower performance when the answer length is 1-2 tokens or 9+ tokens.
This result shows that a question with a short answer is not always an easy one.

\section{Conclusion}
In this paper, we proposed the \textbf{Ja}panese \textbf{Qu}estion \textbf{A}nswering \textbf{D}ataset, JaQuAD.
We collected the contexts from Japanese Wikipedia articles and 39k+ questions were manually annotated by fluent Japanese speakers.
JaQuAD has the same format as SQuAD, and the characteristics of the data are generally similar to KorQuAD 1.0. In the experiments, we fine-tuned a Japanese pre-trained language model with JaQuAD as a baseline and achieved 78.92\% for F1 score and 63.38\% for EM on test set.
The baseline reaches promising results, but there is plenty of room for improvement.
Extension of the dataset, such as covering longer answers, is left for future work.
The dataset and our experiments are available at \texttt{https://github.com/SkelterLabsInc/JaQuAD}.

\section{Acknowledgements}
This work was supported by TPU Research Cloud (TRC) program.
For training models, we used cloud TPUs provided by TRC.
We also thanks to anotators who geernated and labeled JaQuAD.

\bibliographystyle{abbrv}
\bibliography{reference}

\begin{appendices}
\appendix
\section{Criteria for choosing answer spans}
\begin{itemize}
  \item Select the minimum answer span corresponding to the question.
  \item When the answer is proper nouns (e.g. event, book, work name), include parentheses.
  \item When the answer is a year, use basic calendar year and include `年'.
  \item When the answer is numeric, include the basic unit of measurement (e.g. 人, 円, 点, km).
  \item When the answer is an approximate value, include the expressions that indicate approximation (e.g. 約, 以上, 弱).
  \item When the answer is `correct answer (explanation)' form, exclude parentheses except the above cases.
\end{itemize}
\end{appendices}

\end{document}